\documentclass[runningheads]{llncs}
\usepackage{amsmath}
\usepackage{makecell}
\usepackage{booktabs}
\usepackage{arydshln}
\usepackage{multirow}
\usepackage{amsfonts}
\usepackage{subfigure}
\usepackage{cite}
\usepackage[colorlinks,linkcolor=blue,urlcolor=blue,citecolor=blue]{hyperref}
\usepackage{graphicx}
\usepackage{marvosym}

\begin{document}
	
\title{RAGA: Relation-aware Graph Attention Networks for Global Entity Alignment}
\titlerunning{RAGA}
\author{
	Renbo Zhu\inst{1} \and
	Meng Ma\inst{2} \and
	Ping Wang\inst{1,2,3}\textsuperscript{(\Letter)}}
\authorrunning{R. Zhu et al.}
\institute{School of Software and Microelectronics, Peking University, Beijing, China \and
	National Engineering Research Center for Software Engineering, Peking University, Beijing, China \and
	Key Laboratory of High Confidence Software Technologies (PKU), Ministry of Education, Beijing, China\\
	\email{\{zhurenbo,mameng,pwang\}@pku.edu.cn}}
\maketitle

\begin{abstract}
Entity alignment (EA) is the task to discover entities referring to the same real-world object from different knowledge graphs (KGs), which is the most crucial step in integrating multi-source KGs. The majority of the existing embeddings-based entity alignment methods embed entities and relations into a vector space based on relation triples of KGs for local alignment. As these methods insufficiently consider the multiple relations between entities, the structure information of KGs has not been fully leveraged. In this paper, we propose a novel framework based on Relation-aware Graph Attention Networks to capture the interactions between entities and relations. Our framework adopts the self-attention mechanism to spread entity information to the relations and then aggregate relation information back to entities. Furthermore, we propose a global alignment algorithm to make one-to-one entity alignments with a fine-grained similarity matrix. Experiments on three real-world cross-lingual datasets show that our framework outperforms the state-of-the-art methods.
\end{abstract}

\section{Introduction}
Knowledge graphs (KGs) have been widely applied for knowledge-driven artificial intelligence tasks, such as Question Answering~\cite{QA}, Recommendation~\cite{RE} and Knowledge Enhancement~\cite{KE}. 
The completeness of the KGs affects the performance of these tasks. 
Although lots of KGs have been constructed in recent years, none of them can reach perfect coverage due to the defects in the data sources and the inevitable manual process. 
A promising way to increase the completeness of KGs is integrating multi-source KGs, which includes an indispensable step entity alignment (EA). Entity alignment is the task to discover entities referring to the same real-world object from different KGs.

Recently, embedding-based methods have become the dominated approach for entity alignment. They encode entities and relations into a vector space and then find alignments between entities according to their embedding similarities. These methods can be subdivided into two categories: TransE-based methods via translating embeddings (TransE)~\cite{TransE} and GCNs-based methods via graph convolutional networks (GCNs)~\cite{GCNs}. 
However, recent studies point out that there are still the following two critical challenges for entity alignment:

\begin{figure}
	\centering
	\subfigure[An example of multiple relations.]{
		\includegraphics[width=0.65\linewidth]{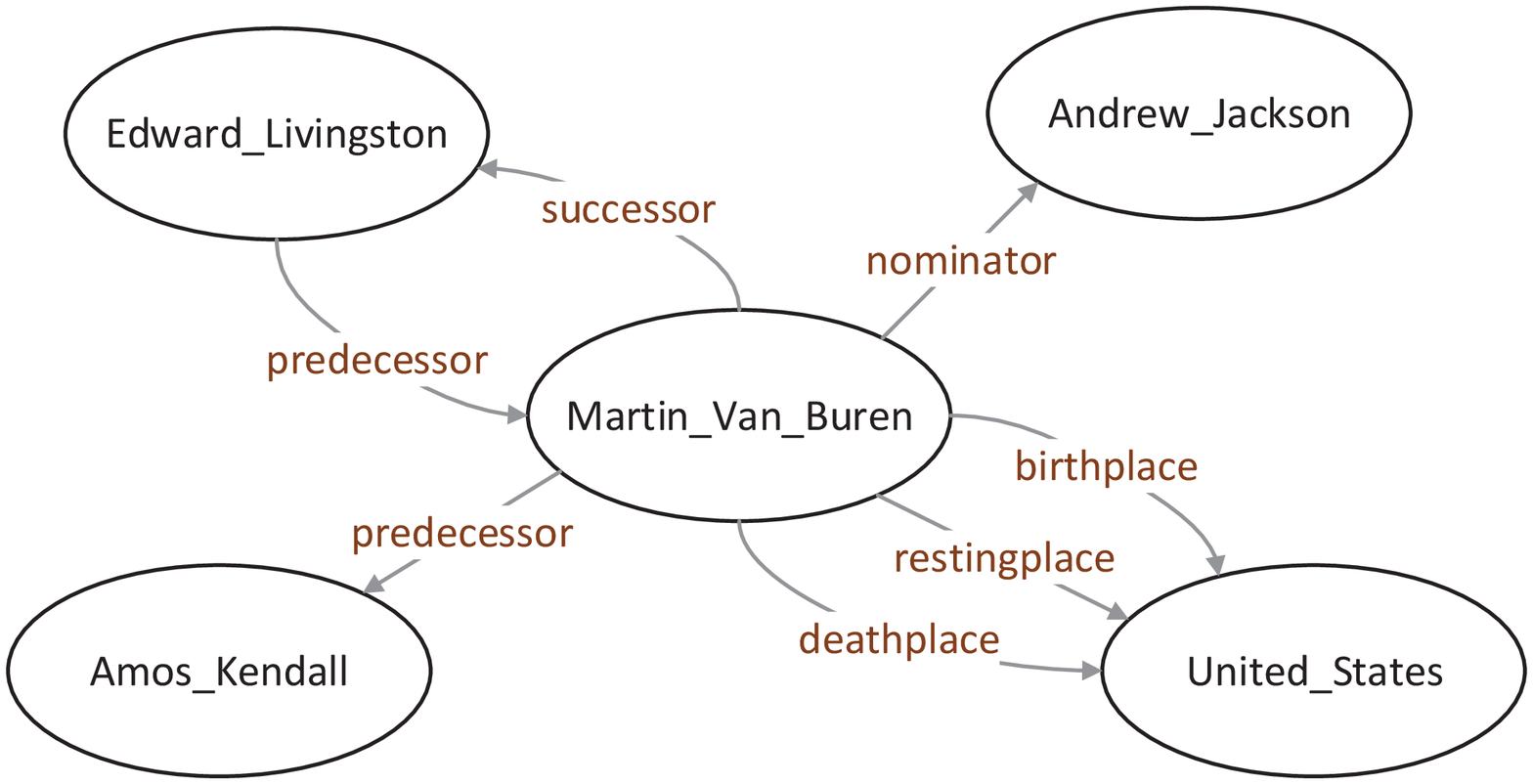}
		\label{relation}
	}
	\subfigure[An example of local alignment conflicts.]{
		\includegraphics[width=0.26\linewidth]{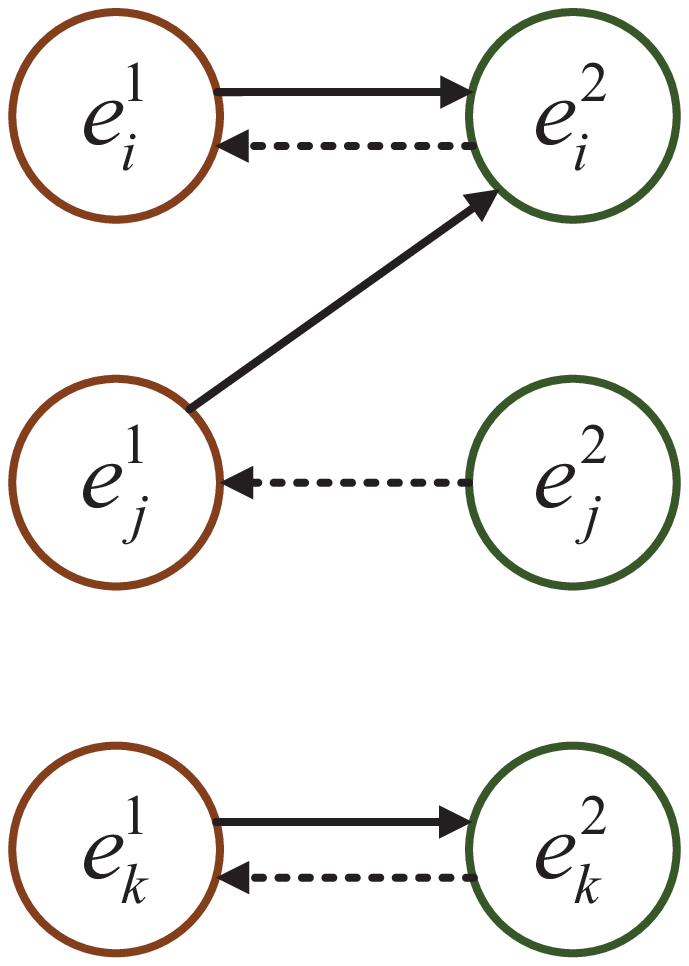}
		\label{conflict}
	}
	\caption{Examples of two challenges for entity alignment.} \label{challenge}
\end{figure}
	\subsubsection{Challenge 1: Sufficient Utilization of Multiple Relations.}
	Figure~\ref{relation} shows a mini KG, in which ellipses represent entities and directed edges represent relations.
	For TransE-based methods, they regard relations as the translation between two entities. However, they are limited by the uniqueness of the relation between two entities. For example, TransE-based methods fail to distinguish \textit{brithplace}, \textit{restingplace} and \textit{deathplace} from \textit{Martin\_Van\_Buren} to \textit{United\_States}. In real life, the above three relations are completely different, and their intersection will contain richer semantic information than any one of them.
	For GCNs-based methods, they model the propagation of entity information based on neighbouring entities on the graph without consideration of corresponding relation types and multiple relations. As an example in the figure, GCNs-based methods spread information of \textit{Martin\_Van\_Buren} to \textit{Edward\_Livingston}, \textit{Andrew\_Jackson}, \textit{Amos\_Kendall} and \textit{United\_States} with equal weights. However, the influence of a person on a person should be different from the influence of a person on a country.
	Thus, the first challenge to entity alignment is how to utilize multiple relations for more reasonable entity representation sufficiently.
	
	\subsubsection{Challenge 2: Global Entity Alignment.}
	Almost all entity alignment methods adopt a local alignment strategy to choose the optimal local match for each entity independently. The local alignment strategy always leads to many-to-one alignment, which means an entity may be the common best match for several entities.
	As an example of alignment results shown in figure~\ref{conflict}, $e^1_i$, $e^1_j$ and $e^1_k$ are entities of $KG_1$,  $e^2_i$, $e^2_j$ and $e^2_k$ are entities of $KG_2$. Solid arrows indicate the best match in $KG_2$ for each entity in $KG_1$, and dotted arrows indicate the best match in $KG_1$ for each entity in $KG_2$. Although $e_k^1$ and $e_k^2$ reach a final match with each other, the best matches of entities $e_i^1$, $e_j^1$, $e_i^2$ and $e_j^2$ lead to conflicts in bidirectional alignment. These conflicts violate the entity alignment task's essential requirement that alignments of two KGs should be interdependent. Thus, the second challenge of entity alignment is how to align entities of two KGs without conflicts from a global perspective.
	
	\subsubsection{Solution.}
	To address above two challenges, we propose a framework RAGA based on \underline{R}elation-aware Graph \underline{A}ttention Networks for \underline{G}lobal Entity \underline{A}lignment. Specially, we propose Relation-aware Graph Attention Networks to capture the interactions between entities and relations, which contributes to sufficient utilization of multiple relations between entities. We then design a global alignment algorithm based on deferred acceptance algorithm, which makes one-to-one entity alignments with a more fine-grained similarity matrix instead of the original embedding similarity matrix. Experimental results on three datasets of cross-lingual KGs demonstrate that RAGA significantly outperforms state-of-the-art baseline methods. The source code is available at \url{https://github.com/zhurboo/RAGA}.

\section{Related Work}
  	\subsection{TransE-Based Entity Alignment}
  	Most of the TransE-based entity alignment methods adopt TransE~\cite{TransE} to learn entity and relation embeddings.
  	With the assumption that the relation is the translation from the head entity to the tail entity in a relation triple, TransE embeds all relations and entities into a unified vector space for a KG.
  	MTransE~\cite{MTransE} encodes entities and relations of each KG in separated embedding space and provides transitions to align the embedding spaces of KGs.
  	JAPE~\cite{JAPE} jointly embeds the structures of two KGs into a unified vector space.
  	TransEdge~\cite{TransEdge} contextualizes relation representations in terms of specific head-tail entity pairs.
  	BootEA~\cite{BootEA} expands seed entity pairs in a bootstrapping way and employs an alignment editing method to reduce error accumulation during iterations.
  	While TransE-based methods can only model fine-grained relation semantics, they cannot preserve the global structure information of KGs with multiple relations.
  	\subsection{GCNs-Based Entity Alignment}
  	With the insight that entities with similar neighbour structures are highly likely to be aligned, GCNs-based entity alignment approaches spread and aggregate entity information on the graph to collect neighbouring entities' representations.
  	GCN-Align~\cite{GCN-Align} is the first attempt to generate entity embeddings by encoding information from their neighbourhoods via GCNs.
  	NMN~\cite{NMN} proposes a neighbourhood matching module with a graph sampling method to effectively construct matching-oriented entity representations.
  	MRAEA~\cite{MRAEA} directly models entity embeddings by attending over the node’s incoming and outgoing neighbours and its connected relations. 
  	RREA~\cite{RREA} leverages relational reflection transformation to obtain relation embeddings for each entity.
  	RDGCN~\cite{RDGCN} incorporates relation information via attentive interactions between the KGs and their dual relation counterpart.
  	HGCN~\cite{HGCN} applies GCNs with Highway Networks gates to embed entities and approximate relation semantics based on entity representations.
  	DGMC~\cite{DGMC} employs synchronous message passing networks to iteratively re-rank the soft correspondences to reach more accurate alignments.
  	Since GCNs has more advantages in dealing with global structure information but ignore local semantic information, MRAEA, RREA, RDGCN, and HGCN make efforts to merge relation information into entity representations. Our framework RAGA adopts a similar idea with more effective interactions between entities and relations.
  	
	\subsection{Global Entity Alignment}
	As each alignment decision highly correlates to the other decisions, every alignment should consider other alignments' influence. Thus, a global alignment strategy is needed for one-to-one alignments. 
	An intuitive idea is calculating the similarity between entities and turning the global entity alignment task into a maximum weighted bipartite matching problem. The Hungarian algorithm~\cite{Hungarian} has been proven to finding the best solution for this problem with the intolerable time complexity of $O(n^4)$ for matching two KGs of $n$ nodes.
	
	To our best knowledge, two methods have been proposed to find an approximate solution for global entity alignment. 
	GM-EHD-JEA~\cite{GM-EHD-JEA} breaks the whole search space into many isolated sub-spaces, where each sub-space contains only a subset of source and target entities for making alignments. It requires a hyper-parameter $\tau$ of the threshold for similarity scores.
  	CEA~\cite{CEA} adopts deferred acceptance algorithm (DAA) to guarantee stable matches. Although CEA achieves satisfactory performance, the similarity matrix it used lacks more fine-grained features, which are considered in our framework RAGA.
  	
\section{Problem Definition}
A KG is formalized as $KG = (E, R, T)$  where $E, R, T$ are the sets of entities, relations and relation triples, respectively. A relation triple  $(h,r,t)$ consists of a head entity $h\in E$, a relation $r \in R$ and a tail entity $t\in E$.

Given two KGs, $KG_1=(E_1,R_1,T_1)$ and $KG_2=(E_2,R_2,T_2)$, we define the task of entity alignment as discovering equivalent entities based on a set of seed entity pairs as $S = \{(e_1, e_2)|e_1 \in E_1, e_2 \in E_2, e_1 \leftrightarrow e_2\}$, where $\leftrightarrow$ represents equivalence.

For global entity alignment, it requires one-to-one matches, which means that each entity to be aligned has its equivalent entity, and the entity alignment results do not contain any alignment conflicts.
\section{RAGA Framework}
\begin{figure}
	\includegraphics[width=\textwidth]{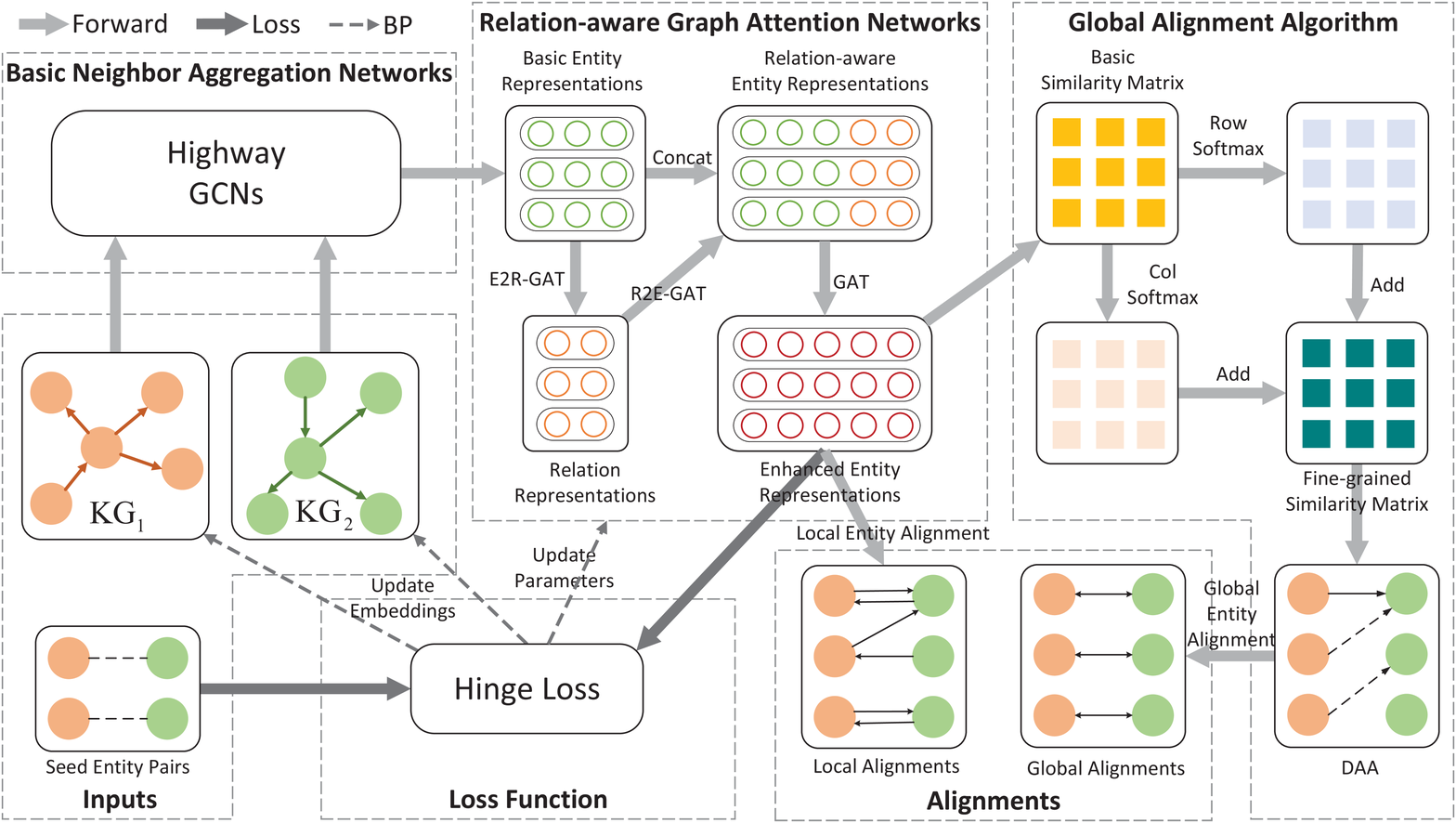}
	\caption{Overall architecture of RAGA framework.} \label{framework}
\end{figure}

We propose our RAGA framework based on interactions between entities and relations via the self-attention mechanism. Figure~\ref{framework} depicts the overall architecture of our framework, which mainly consists of four parts: Basic Neighbor Aggregation Networks, Relation-aware Graph Attention Networks, End-to-End Training and Global Alignment Algorithm. First, we adopt Basic Neighbor Aggregation Networks to obtain basic entity representations. Then, we generate enhanced entity representations via Relation-aware Graph Attention Networks, which incorporates relation information into entities. In the End-to-End Training part, the embeddings of input entities and the parameters of Relation-aware Graph Attention Networks are updated via backpropagation. Finally, the global alignment algorithm is applied to generate global alignments.

	\subsection{Basic Neighbor Aggregation Networks}
	To get basic entity representations, we utilize GCNs to explicitly encode entities in KGs with structure information. The input of $l$-th GCN layer is a set of entity embeddings $\boldsymbol{X}^{(l)}=\left\{\boldsymbol{x}_{1}^{(l)}, \boldsymbol{x}_{2}^{(l)}, \cdots, \boldsymbol{x}_{n}^{(l)} \mid \boldsymbol{x}_{i}^{(l)} \in \mathbb{R}^{d_e}\right\}$, where $n$ is the number of entities, and $d_e$ is the dimension of entity embeddings, the output of the $l$-th layer is obtained following the convolution computation:
	\begin{equation}
	{\boldsymbol{X}}^{(l+1)}=\operatorname{ReLU}\left(\tilde{D}^{-\frac{1}{2}} \tilde{A} \tilde{D}^{-\frac{1}{2}} {\boldsymbol{X}}^{(l)} \right),
	\end{equation}
	where $\tilde{A}=A+I$, $A$ is the adjacency matrix of $KG$, $I$ is an identity matrix, and $\tilde{D}$ is the diagonal node degree matrix of $\tilde{A}$. As entity embeddings are learnable, we do not apply a trainable weight matrix to change the distribution of ${\boldsymbol{X}}^{(l)}$, which may lead to overfitting.
	
	Inspired by RDGCN~\cite{RDGCN}, we employ
	layer-wise Highway Networks~\cite{Highway} to control the balance of the information between the entity itself and neighbour entities. The output of a Highway Network layer is the weighted sum of its input and the original output via gating weights:
	\begin{equation}
	T\left(\boldsymbol{X}^{(l)}\right)=\sigma\left(\boldsymbol{X}^{(l)} \boldsymbol{W}^{(l)}+\boldsymbol{b}^{(l)}\right),
	\end{equation}
	\begin{equation}
	\boldsymbol{X}^{(l+1)}=T\left(\boldsymbol{X}^{(l)}\right) \cdot \boldsymbol{X}^{(l+1)}+\left(1-T\left(\boldsymbol{X}^{(l)}\right)\right) \cdot \boldsymbol{X}^{(l)},
	\end{equation}
	where $\sigma$ is a sigmoid function, $\cdot$ is element-wise multiplication, $\boldsymbol{W}^{(l)}$ and $\boldsymbol{b}^{(l)}$ are the weight matrix and bias vector
	for the transform gate of the $l$-th layer.
	\subsection{Relation-aware Graph Attention Networks}
	To obtain more accurate entity representations, we propose Relation-aware Graph Attention Networks, which sequentially pass the entity representations through the three diffusion modes of entity to relation, relation to entity, and entity to entity. We use a similar format to describe the above three diffusion modes.
		\subsubsection{Relation Representations.}
		As the distribution of relations is denser than the distribution of entities, relation representations should have a different amount of information and different information space from entity representations.
		Thus, we apply a linear transition to entity embeddings, then calculate relation embeddings with attention weights. For each relation, we leverage its connected head entities and tail entities, which will be embedded in two vectors through their respective linear transition matrices. Different from RDGCN~\cite{RDGCN}, our relation representations do not ignore the duplicate links between entities and relations, which is used to adjust the attention weights. 
		
		For relation $r_k$, the head entity representation $\boldsymbol{r}^h_k$ is computed as follows:
		\begin{equation}
		\alpha_{ijk} = \frac{\operatorname{exp}\left(\operatorname{LeakReLU}\left(\boldsymbol{a}^T\left[\boldsymbol{x}_{i}\boldsymbol{W}^h \| \boldsymbol{x}_j\boldsymbol{W}^t \right]\right)\right)}
		{\sum_{e_{i^\prime} \in \mathcal{H}_{r_{k}}} \sum_{e_{j^\prime} \in \mathcal{T}_{e_i r_k}}\operatorname{exp}\left(\operatorname{LeakReLU}\left(\boldsymbol{a}^T\left[\boldsymbol{x}_{i^\prime}\boldsymbol{W}^h \| \boldsymbol{x}_{j^\prime}\boldsymbol{W}^t\right]\right)\right)},
		\end{equation}
		\begin{equation}
		\boldsymbol{r}_{k}^{h}=\operatorname{ReLU}\left(\sum_{e_{i} \in \mathcal{H}_{r_{k}}} \sum_{e_{j} \in \mathcal{T}_{e_i r_k}} \alpha_{ijk} \boldsymbol{x}_{i}\boldsymbol{W}^h\right),
		\end{equation}
		where $\alpha_{ijk}$ represents attention weight from head entity $e_i$ to relation $r_k$ based on head entity $e_i$ and tail entity $e_j$, $\mathcal{H}_{r_{k}}$ is the set of head entities for relation $r_k$, $\mathcal{T}_{e_i r_k}$ is the set of tail entities for head entity $e_i$ and relation $r_k$, $\boldsymbol{a}$ is a one-dimensional vector to map the $2d_r$-dimensional input into a scalar, $d_r$ is half of the dimension of relation embeddings, and $\boldsymbol{W}^h, \boldsymbol{W}^t \in \mathbb{R}^{d_e\times d_r}$ are linear transition matrices for head and tail entity representation of relations respectively. 
		
		We can compute the tail entity representation $\boldsymbol{r}^t_k$ through a similar process, and then add them together to obtain the relation representation $\boldsymbol{r}_k$:
		\begin{equation}
		\boldsymbol{r}_{k}=\boldsymbol{r}_{k}^{h}+\boldsymbol{r}_{k}^{t}.
		\end{equation}

		\subsubsection{Relation-aware Entity Representations.}
		Based on the experience that an entity with its neighbour relations is more accurately expressing itself, we regroup the embeddings of relation adjacents into the entity representations. 
		Specifically, for entity $e_i$, we adopt attention mechanism to calculate its out-relation ($e_i$ is the head of those relations) embedding $\boldsymbol{x}_i^h$ and in-relation ($e_i$ is the tail of those relations) embedding $\boldsymbol{x}_i^t$ separately. $\boldsymbol{x}_i^h$ is computed as follows:
		\begin{equation}
		\alpha_{ik} = \frac{\operatorname{exp}\left(\operatorname{LeakReLU}\left(\boldsymbol{a}^T\left[\boldsymbol{x}_i \|\boldsymbol{r}_k \right]\right)\right)}
		{\sum_{e_{j} \in \mathcal{T}_{e_{i}}} \sum_{r_{k^\prime} \in \mathcal{R}_{e_i e_j}}\operatorname{exp}\left(\operatorname{LeakReLU}\left(\boldsymbol{a}^T\left[\boldsymbol{x}_i \|\boldsymbol{r}_{k^\prime} \right]\right)\right)},
		\end{equation}
		\begin{equation}
		\boldsymbol{x}_{i}^{h}=\operatorname{ReLU}\left(\sum_{e_{j} \in \mathcal{T}_{e_{i}}} \sum_{r_{k} \in \mathcal{R}_{e_i e_j}} \alpha_{i k} \boldsymbol{r}_{k}\right),
		\end{equation}
		where $\alpha_{ik}$ represents attention weight from relation $r_k$ to entity $e_i$, $\mathcal{T}_{e_{i}}$ is the set of tail entities for head entity $e_i$ and $\mathcal{R}_{e_i e_j}$ is the set of relations between head entity $e_i$ and tail entity $e_j$.
		
		Then the relation-aware entity representations $\boldsymbol{x}_{i}^{rel}$ can be expressed by concatenating $\boldsymbol{x}_{i}$, $\boldsymbol{x}_{i}^{h}$ and $\boldsymbol{x}_{i}^{t}$:
		\begin{equation}
		\boldsymbol{x}_{i}^{rel}=\left[\boldsymbol{x}_{i}\|\boldsymbol{x}_{i}^{h}\|\boldsymbol{x}_{i}^{t}\right].
		\end{equation}
		\subsubsection{Enhanced Entity Representations.}
		In relation-aware entity representations, entities only contain the information of one-hop relations. To enhance the influence of relations on two-hop entities, we adopt one layer of ordinary graph attention networks to get enhanced entity representations. This process considers bidirectional edges and does not include a linear transition matrix.
		For entity $e_i$, the final output of embedding $\boldsymbol{x}_i^{out}$ can be computed by:
		\begin{equation}
		\alpha_{i j}=\frac{\exp \left(\operatorname{LeakyReLU}\left(\boldsymbol{a}^T\left[\boldsymbol{x}_{i}^{rel} \| \boldsymbol{x}_{j}^{rel}\right]\right)\right)}{\sum_{j^\prime \in \mathcal{N}_{i}} \exp \left(\text {LeakyReLU}\left(\boldsymbol{a}^T\left[\boldsymbol{x}_{i}^{rel} \| \boldsymbol{x}_{j^\prime}^{rel}\right]\right)\right)},
		\end{equation}
		
		\begin{equation}
		\boldsymbol{x}_{i}^{out}=\left[\boldsymbol{x}_{i}^{rel} \| \operatorname{ReLU}\left(\sum_{j \in \mathcal{N}_{i}} \alpha_{i j}\boldsymbol{x}_{i}^{rel}\right)\right].
		\end{equation}
	
	\subsection{End-to-End Training}
	We use Manhattan distance to calculate the similarity of entities:
	\begin{equation}
	\operatorname{dis}\left(e_{i}, e_{j}\right)=\left\|\boldsymbol{x}_{{i}}^{out}-\boldsymbol{x}_{{j}}^{out}\right\|_{1}.
	\end{equation}
	
	For ent-to-end training, we regard all relation triples $T$ in KGs as positive samples. Every $p$ epoch, we adopt the nearest neighbour sampling to sample $k$ negative samples from each knowledge graph for each entity.
	Finally, we use Hinge Loss as our loss function:
	\begin{equation}
	L=\sum_{\left(e_{i}, e_{j}\right) \in T}\sum_{\left(e^\prime_{i}, e^\prime_{j}\right) \in T^\prime_{(e_i,e_j)}}\max \left(\operatorname{dis}\left(e_{i}, e_{j}\right)-\operatorname{dis}\left(e_{i}^{\prime}, e_{j}^{\prime}\right)+\lambda, 0\right),
	\end{equation}
	where $T^\prime_{(e_i,e_j)}$ is the set of negative sample for $e_i$ and $e_j$, $\lambda$ is margin.
	
	\subsection{Global Alignment Algorithm}
	As optimal local matches for entity alignment may lead to many-to-one alignments that reduce performance and bring ambiguity to entity alignment results, entities should be aligned globally. Thus, we design a global alignment algorithm.
	
	Through Basic Neighbor Aggregation Networks and Relation-aware Graph Attention Networks, we obtain entity embeddings for each entity of two KGs. Then a similarity matrix $S \in \mathbb{R}^{|E_1|\times |E_2|}$ can be constructed based on the Manhattan distance between every two entity from different KGs. While CEA ~\cite{CEA} directly applied deferred acceptance algorithm (DAA)~\cite{DAA} to the similarity matrix $S$ for global entity alignment, we argue that more fine-grained features can be merged into the matrix. According to prior knowledge, entity alignment is a bidirectional match problem between two KGs. Thus, we calculate a fine-grained similarity matrix $S^g$ by summing the weights of each entity aligned in two directions. Specifically, we adopt softmax on both rows and columns of $S$ and add them together to get the fine-grained similarity matrix $S^g$:
	\begin{equation}
	S^{g}_{i,j} = \frac{\mathrm{exp}(S_{i,j})}{\sum_{j^\prime=1}^{|E_2|}\mathrm{exp}(S_{i,j^\prime})}+\frac{\mathrm{exp}(S_{i,j})}{\sum_{i^\prime=1}^{|E_1|}\mathrm{exp}(S_{i^\prime,j})}.
	\end{equation}
	
	Finally, we also adopt DAA to the fine-grained similarity matrix $S^g$ to get global alignments. The detailed process of DAA can refer to~\cite{CEA}. The time complexity of the  alignment process is $O(|E_1|\cdot|E_2|\cdot {\rm log}(|E_1|\cdot|E_2|))$, which is much smaller than that of  Hungarian algorithm.

\section{Experiments}
	\subsection{Experimental Settings}
	    \subsubsection{Dataset.}
		We evaluate the proposed framework on DBP15K~\cite{JAPE}. It contains three pairs of cross-lingual KGs: ZH-EN, JA-EN, and FR-EN. Each dataset includes 15,000 alignment entity pairs. Almost all entity alignment studies based on DBP15K adopt a simplified version of DBP15K, which removes lots of unrelated entities and relations. Our experiment is also based on the simplified version of DBP15K, which is shown in the table~\ref{dataset}. For each dataset, we use 30\% of the alignment entity pairs as seed entity pairs for training and 70\% for testing.
	    \begin{table}
	    	\caption{Statistical data of simplified DBP15K.}\label{dataset}
	    	\centering
	    	\begin{tabular}{p{30pt}<{\centering}p{20pt}<{\centering}p{50pt}<{\centering}p{50pt}<{\centering}p{60pt}<{\centering}p{72pt}<{\centering}}
	    		\toprule[1pt]
	    		\multicolumn{2}{c}{DBP15K}  & \#Entities & \#Relations & \#Rel Triples & \#Ent Alignments        \\
	    		\midrule[1pt]
	    		\multirow{2}{*}{ZH-EN} & ZH & 19,388     & 1,700       & 70,414        & \multirow{2}{*}{15,000} \\
	    		& EN & 19,572     & 1,322       & 95,142        &                         \\
	    		\hline
	    		\multirow{2}{*}{JA-EN} & JA & 19,814     & 1,298       & 77,214        & \multirow{2}{*}{15,000} \\
	    		& EN & 19,780     & 1,152       & 93,484        &                         \\
	    		\hline
	    		\multirow{2}{*}{FR-EN} & FR & 19,661     & 902       & 105,998       & \multirow{2}{*}{15,000} \\
	    		& EN & 19,993     & 1,207       & 115,722       &    \\
	    		\bottomrule[1pt]                    
	    	\end{tabular}
	    \end{table}
	    \subsubsection{Baselines.}
	    To comprehensively evaluate our framework, we compare to both TransE-based, GCNs-based and global entity alignment methods: 
	    \begin{itemize}
	    \item TransE-based methods: MtransE~\cite{MTransE}, JAPE~\cite{JAPE}, BootEA~\cite{BootEA}, TransEdge~\cite{TransEdge}.
	    \item GCNs-based methods: GCN-Align~\cite{GCN-Align}, MRAEA~\cite{MRAEA}, RREA~\cite{RREA}, RDGCN~\cite{RDGCN}, HGCN~\cite{HGCN}, NMN~\cite{NMN}, DGMC~\cite{DGMC}.
	    \item Global methods: GM-EHD-JEA~\cite{GM-EHD-JEA}, CEA~\cite{CEA}.
	    \end{itemize}
		For a fair comparison, we do not compare with methods that require additional information, such as entity descriptions, attributes and attribute values.
		To our best knowledge, DGMC and CEA are the state-of-the-art methods for local and global entity alignment respectively without additional information.
	    \subsubsection{Evaluation Metrics.}
	    For local entity alignment, followed~\cite{MTransE}, we use Hitratio@K (H@k) and mean reciprocal rank (MRR) to measure the performance.
	    For global entity alignment, since one-to-one alignment results are produced, only H@1 was adopted. 
	    For all metrics, the larger, the better.
	    \subsubsection{Implementation Details.}
	    Following~\cite{RDGCN}, we translate all entity names to English via Google Translate and then use Glove~\cite{Glove} to construct the initial entity embeddings. On each language pair in DBP15k, we randomly divide alignment pairs, 30\% for training and 70\% for testing, which is the same as previous works. In Basic Neighbor Aggregation Networks, the depth of Highway-GCNs $l$ is 2. In Relation-aware Graph Attention Networks, the half of dimension of relation embeddings $d_r$ is 100. For end-to-end training, the number of epochs for updating negative samples $p$ is 5, and the negative sample number $k$ is 5. In margin-based loss function, the margin $\lambda$ is 3.0.
	    \subsubsection{Model Variants.}
	    In order to study the effectiveness of each component in our framework, we provide the following different variants of RAGA: 
	    \begin{itemize}
	   	\item Init-Emb: The initial entity embeddings, which are also applied in RDGCN, HGCN, NMN, DGMC, and CEA.
	   	\item w/o RGAT: Our framework without Relation-aware Graph Attention Networks for local entity alignment.
	   	\item w/o BNA: Our framework without Basic Neighbor Aggregation Networks for local entity alignment.
	   	\item RAGA-l: Our framework for local entity alignment.
	   	\item w/o Bi: Our framework without the fine-grained similarity matrix.
	    \end{itemize}
	\subsection{Experimental Results and Analysis}	
	Table~\ref{result} shows the overall results of all methods. All comparable results except DGMC are taken from their original papers. Since the original paper of DGMC uses the different version of DBP15K, we run the source code with the same dataset we use to get its results. The parts of the results separated by solid line denote TransE-based methods, GCNs-based methods and global methods. The last parts below the dashed line of GCNs-based methods and global methods are the results of our models.
	\begin{table}
		\caption{Overall performance of entity alignment.}\label{result}
		\centering
		\begin{tabular}{p{65pt}<{\centering}p{28pt}<{\centering}p{28pt}<{\centering}p{28pt}<{\centering}p{28pt}<{\centering}p{28pt}<{\centering}p{28pt}<{\centering}p{28pt}<{\centering}p{28pt}<{\centering}p{28pt}<{\centering}}
			\toprule[1pt]
			& \multicolumn{3}{c}{ZH-EN} & \multicolumn{3}{c}{JA-EN} & \multicolumn{3}{c}{FR-EN} \\ 
			Methods       & H@1    & H@10   & MRR     & H@1    & H@10   & MRR     & H@1    & H@10   & MRR     \\ 
			\midrule[1pt]
			MTransE       & 30.8   & 61.4   & 0.364   & 27.9   & 57.5   & 0.349   & 24.4   & 55.6   & 0.335   \\
			JAPE          & 41.2   & 74.5   & 0.490   & 36.3   & 68.5   & 0.476   & 32.3   & 66.7   & 0.430   \\
			BootEA        & 62.9   & 84.8   & 0.703   & 62.2   & 85.4   & 0.701   & 65.3   & 87.4   & 0.731   \\
			TransEdge     & 73.5   & 91.9   & 0.801   & 71.9   & 93.2   & 0.795   & 71.0   & 94.1   & 0.796   \\ 
			\hline
			GCN-Align     & 41.3   & 74.4   & 0.549   & 39.9   & 74.5   & 0.546   & 37.3   & 74.5   & 0.532   \\
			MRAEA         & 63.5   & 88.2   & 0.729   & 63.6   & 88.7   & 0.731   & 66.6   & 91.2   & 0.764   \\
			RREA          & 71.5   & 92.9   & 0.794   & 71.3   & 93.3   & 0.793   & 73.9   & 94.6   & 0.816   \\
			RDGCN         & 70.8   & 84.6   & -       & 76.7   & 89.5   & -       & 88.6   & 95.7   & -       \\
			HGCN          & 72.0   & 85.7   & -       & 76.6   & 89.7   & -       & 89.2   & 96.1   & -       \\
			NMN           & 73.3   & 86.9   & -       & 78.5   & 91.2   & -       & 90.2   & 96.7   & -       \\
			DGMC          & 74.8   & 82.5   & -       & 80.4   & 86.4   & -       & \textbf{93.1}   & 95.8   & -       \\
			\hdashline
			Init-Emb      & 57.5   & 68.9   & 0.615   & 65.0   & 75.4   & 0.688   & 81.8   & 88.8   & 0.843   \\
			w/o RGAT   & 74.7   & 86.4   & 0.790   & 78.5   & 89.6   & 0.826   & 89.9   & 96.0   & 0.922   \\
			w/o BNA  & 76.0   & 88.1   & 0.805   & 79.5   & 89.8   & 0.833   & 90.9   & 96.3   & 0.930   \\
			RAGA-l       & \textbf{79.8}   & \textbf{93.0}   & \textbf{0.847}   & \textbf{83.1}   & \textbf{95.0}   & \textbf{0.875}   & 91.4   & \textbf{98.3}   & \textbf{0.940}   \\ 
			\hline
			GM-EHD-JEA    & 73.6   & -      & -       & 79.2   & -      & -       & 92.4   & -      & -       \\
			CEA           & 78.7   & -      & -       & 86.3   & -      & -       & \textbf{97.2}   & -      & -       \\
			\hdashline
			w/o Bi      & 84.3   & -      & -       & 86.7   & -      & -       & 94.1   & -      & -       \\
			RAGA      & \textbf{87.3}   & -      & -       & \textbf{90.9}   & -      & -       & 96.6   & -      & -       \\ 
			\bottomrule[1pt]
		\end{tabular}
	\end{table}
		\subsubsection{Overall EA Performance.}
		For TransE-based methods, BootEA and TransEdge outperform MTransE and JAPE with their iterative strategies. Furthermore, by contextualizing relation representations in terms of specific head-tail entity pairs and interpreting them as translations between entity embeddings, TransEdge achieves excellent performance with random initial entity embeddings.	
		
		For GCNs-based methods, GCN-Align performs worst due to simple utilization relation triples. As MRAEA and RREA leverage more relation information, they get much better performance than GCN-Align. Based on initial entity embeddings, RDGCN, HGCN, NMN, and DGMC further improve their performance. Combining Basic Neighbor Aggregation Networks and Relation-aware Graph Attention Networks, our RAGA-l performs best in almost all evaluation metrics. It is noteworthy that DGMC performs 1.7\% better than RAGA-l on H@1 of FR-EN. The reasons can be summarized as the following two points. First, in FR-EN dataset, due to high language similarity, the init embeddings contain rich information, which reduces the difficulty of the alignment task. Second, DGMC employs synchronous message passing networks, which is conducive to close H@1 to H@10. Thus, with a small gap of H@10 between DGMC and RAGA-l, DGMC has an advantage on H@1.
	
		For global entity alignment methods, combined with the global alignment algorithm, our RAGA outperforms other methods in ZH-EN and JA-EN datasets. CEA performs slightly better than our RAGA in FR-EN. It is because CEA leverages extra entity descriptions, which are not considered in our methods.
		\subsubsection{Effect of Relation-aware Graph Attention Networks.}
		To analyze the effect of Relation-aware Graph Attention Networks, we construct three variants of RAGA-l: Init-Emb, w/o RGAT and w/o BNA. 
		From the results, we can see that while both Basic Neighbor Aggregation Networks and Relation-aware Graph Attention Networks improve the performance significantly. Relation-aware Graph Attention Networks has a more significant effect than the former.
		\subsubsection{Effect of Global alignment Algorithm.}
		To analyze the effect of our global alignment algorithm, we compare RAGA with w/o Bi, which adopts the same global alignment strategy as CEA. Experiments show that our global alignment algorithm with the fine-grained similarity matrix $S^g$ further brings 2.5-4.3\% improvement based on good enough alignment results.
		\subsubsection{Impact of Seed Entity Pairs.}
\begin{figure}
		\centering
		\subfigure[ZH-EN]{
			\includegraphics[width=0.31\linewidth]{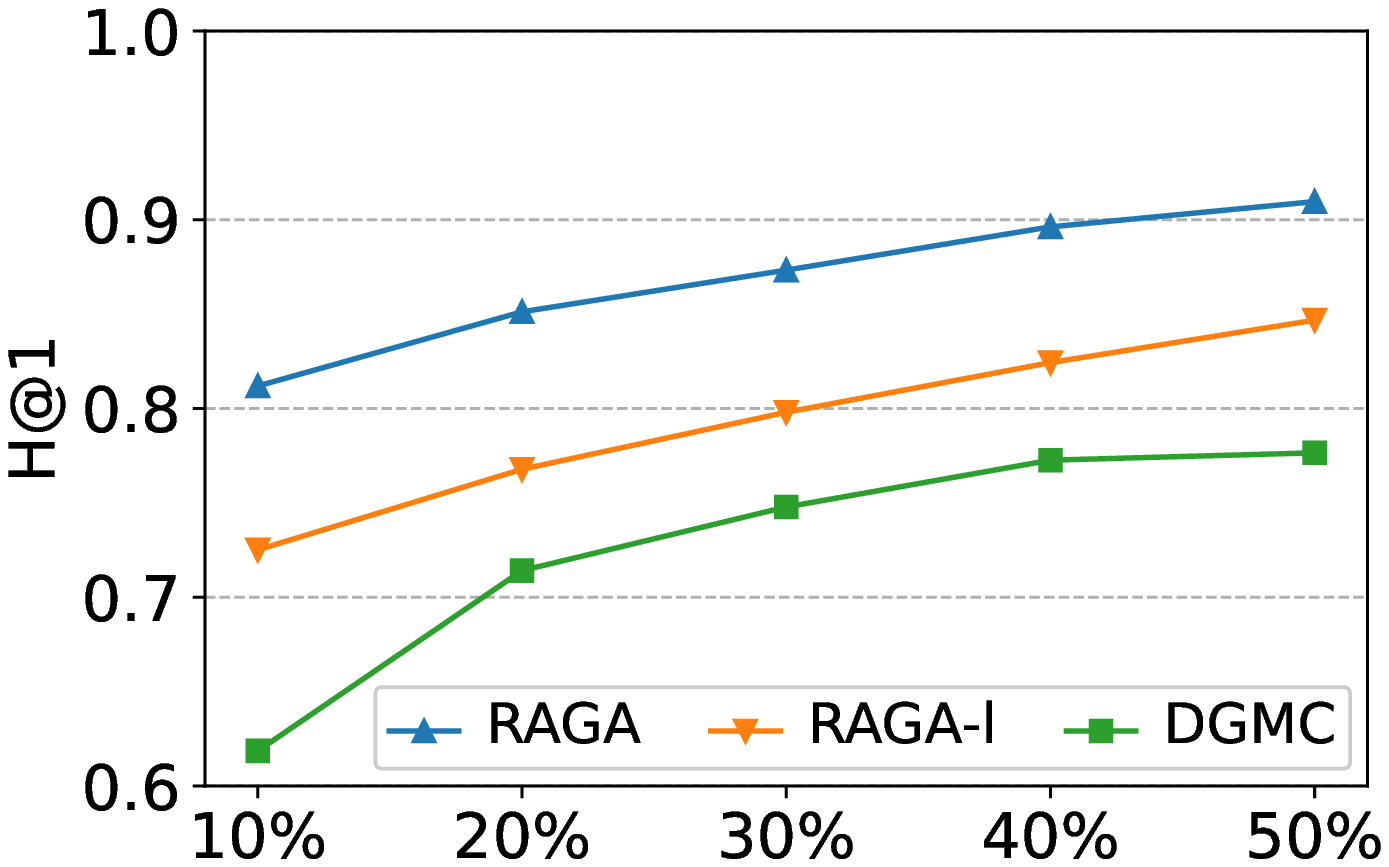}
		}
		\subfigure[JA-EN]{
			\includegraphics[width=0.31\linewidth]{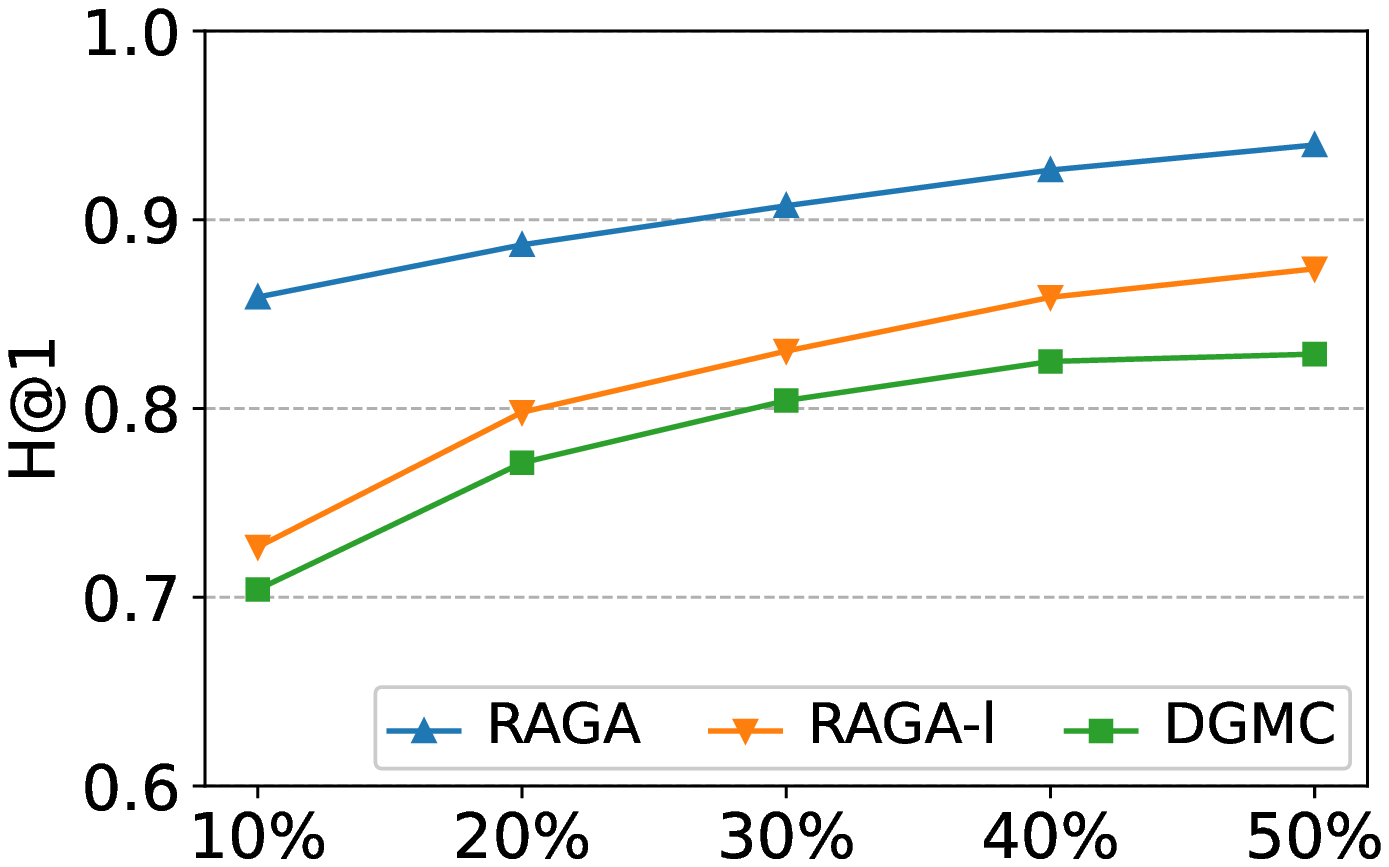}
		}
		\subfigure[FR-EN]{
			\includegraphics[width=0.31\linewidth]{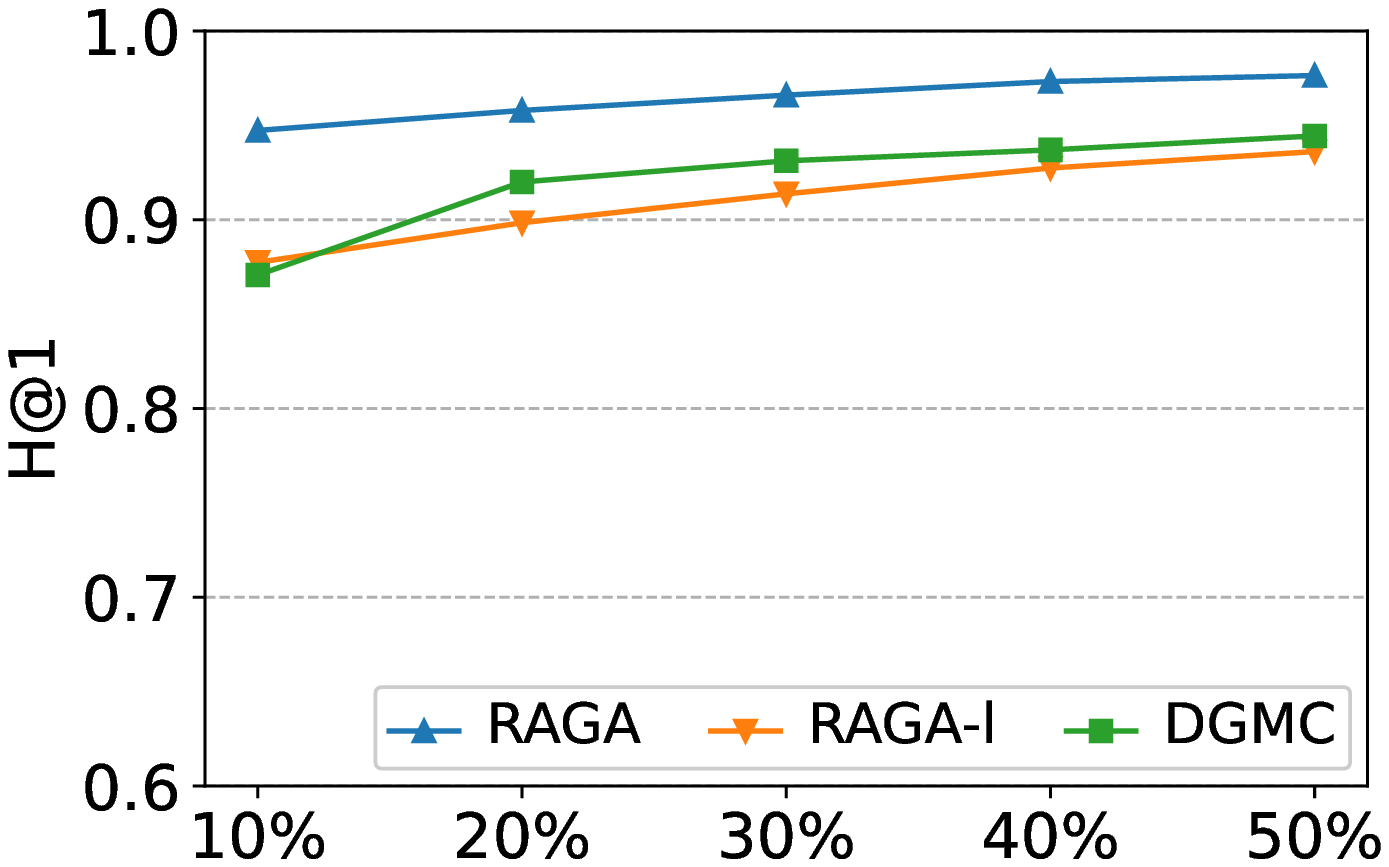}
		}
		\caption{H@1 of entity alignment results with different seed entity pairs.} \label{rate}
		\end{figure}

		To explore the impact of seed entity pairs on our framework, we compare RAGA and RAGA-l with DGMC by varying the proportion of seed entity pairs from 10\% to 50\% with a step size of 10\%. Figure~\ref{rate} depicts H@1 with respect to different proportions. It seems that when seed entity pairs increase, RAGA and RAGA-l have more room for improvement while the performance of DGMC gradually reaches the bottleneck. Moreover, RAGA has a more gradual slope curve, which means the good capability of generalization.
\section{Conclusion}
In this paper, we have investigated the problem of entity alignment for the fusion of KGs. To address sufficient utilization of multiple relations and global entity alignment, we propose our framework RAGA to model the interactions between entities and relations for global entity alignment. Combined with Relation-aware Graph Attention Networks and global alignment algorithm, our framework outperforms the state-of-the-art entity alignment methods on three real-world cross-lingual datasets.

\subsubsection{Acknowledgement.}
This work is supported by National Key Research and Development Program of China under Grant 2017YFB1200700.

\end{document}